\documentclass[sigconf]{acmart}
\usepackage{xcolor}

\usepackage{tabularx} 
\usepackage{listings}
\usepackage{subcaption}

\usepackage[breakable]{tcolorbox}
\usepackage{subcaption}
\usepackage{algorithm}
\usepackage[noend]{algpseudocode}
\tcbuselibrary{most}
\algnewcommand{\LineComment}[1]{\State \(\triangleright\) \textit{#1}}
\algnewcommand{\Continue}{\State\textbf{continue}}
\usepackage{listings}
\usepackage{xspace} 
\newcommand\st{\textsuperscript{st}\xspace}
\newcommand\nd{\textsuperscript{nd}\xspace}
\newcommand\rd{\textsuperscript{rd}\xspace}

\setcopyright{none}

\AtBeginDocument{%
  }

\begin{document}

\title{SafePilot: A Framework for Assuring LLM-enabled Cyber-Physical Systems}



\author{Weizhe Xu}
\affiliation{%
  \institution{University of Notre Dame}
  \city{Notre Dame, IN}
  \country{USA}
}
\email{wxu3@nd.edu}

\author{Mengyu Liu}
\affiliation{%
  \institution{Washington State University}
  \city{Richland, WA}
  \country{USA}
}
\email{mengyu.liu@wsu.edu}

\author{Fanxin Kong}
\affiliation{%
  \institution{University of Notre Dame}
  \city{Notre Dame, IN}
  \country{USA}
}
\email{fkong@nd.edu}

\begin{abstract}
Large Language Models (LLMs), deep learning architectures with typically over 10 billion parameters, have recently begun to be integrated into various cyber-physical systems (CPS) such as robotics, industrial automation, and autopilot systems. The abstract knowledge and reasoning capabilities of LLMs are employed for tasks like planning and navigation. However, a significant challenge arises from the tendency of LLMs to produce "hallucinations"—outputs that are coherent yet factually incorrect or contextually unsuitable. This characteristic can lead to undesirable or unsafe actions in the CPS. Therefore, our research focuses on assuring the LLM-enabled CPS by enhancing their critical properties. We propose SafePilot, a novel hierarchical neuro-symbolic framework that provides end-to-end assurance for LLM-enabled CPS according to attribute-based and temporal specifications. Given a task and its specification, SafePilot first invokes a hierarchical planner with a discriminator that assesses task complexity. If the task is deemed manageable, it is passed directly to an LLM-based task planner with built-in verification. Otherwise, the hierarchical planner applies a divide-and-conquer strategy, decomposing the task into sub-tasks, each of which is individually planned and later merged into a final solution. The LLM-based task planner translates natural language constraints into formal specifications and verifies the LLM's output against them. If violations are detected, it identifies the flaw, adjusts the prompt accordingly, and re-invokes the LLM. This iterative process continues until a valid plan is produced or a predefined limit is reached. 
Our framework supports LLM-enabled CPS with both attribute-based and temporal constraints. Its effectiveness and adaptability are demonstrated through two illustrative case studies.

\end{abstract}

\keywords{large language models, cyber-physical systems, assurance, formal language, formal verification}

\maketitle

\section{Introduction}
Cyber-physical systems (CPS)~\cite{choi2018detecting, dan2010stealth, pasqualetti2011cyber, liu2024cpsim, liu2022fail} integrate computing and networking components to control the physical system and interact with the environment using sensors and actuators. 
Distinguished by their autonomous and adaptive features, CPS show a remarkable advancement beyond traditional control systems.
To further enhance their intelligence and versatility, researchers have continuously explored new paradigms.
In parallel, Large Language Models (LLMs), such as GPT-4~\cite{sanderson2023gpt} and LLeMA~\cite{touvron2023llama} have achieved remarkable progress, demonstrating strong abilities, such as perception, interaction, decision-making and reasoning~\cite{xu2024llm}.
Recent studies have begun to investigate the integration of LLMs into CPS to provide sophisticated intelligence support.
In some scenarios, LLMs act as assistants in tasks such as data processing, contextual understanding, and information retrieval~\cite{brohan2023rt, fu2024drive}.
In other cases, LLMs serve as decision-makers, generating task plans and guiding CPS to accomplish specific objectives~\cite{cui2023large, lin2023text2motion, singh2023progprompt, inagaki2023llms, nascimento2023gpt}.
These LLM-enabled CPS that employ LLMs as decision-makers are the primary focus of this study.

While LLMs significantly enhance the intelligence of CPS by providing advanced reasoning capabilities, the probabilistic nature of LLMs conflicts with the deterministic behavior expected in CPS.
This conflict introduces risks and instability, potentially leading to severe consequences~\cite{chen2023can}.
For instance, a robot following an LLM-generated plan could collide with obstacles or miss the deadline.
This issue, called hallucination, happens when the model generates responses that seem reasonable but are actually incorrect, meaningless, or made up.
Hallucinations can arise from various sources and are often unavoidable, including the absence of sufficient contextual information and the inherent limitations in the reasoning capabilities of LLMs.
This hallucination phenomenon in LLMs poses significant risks to LLM-enabled CPS, severely undermining their adoption and deployment in real-world applications.
Therefore, assuring the reliability and safety of these LLM-enabled CPS has become an urgent and critical problem that must be addressed.

Although various testing and assurance techniques have been proposed for both neural networks and CPS, none of them can effectively assure the reliability of LLM-enabled CPS.
As for assuring neural networks~\cite{zheng2023testing, katz2017reluplex, liu2021algorithms}, these approaches are inadequate for LLMs, due to the enormous number of parameters and the wide range of their applications.
Furthermore, traditional verification and safety assurance approaches for CPS are inadequate for handling the adaptive and evolving nature of LLMs. 
In addition, users seek to not only verify the outputs of LLMs but also guide their behavior to ensure accurate and reliable results. 
This paper aims to present a robust and generalizable framework specifically designed for LLM-enabled CPS, which not only assures critical specifications such as safety but also enhances the planning capabilities of LLMs.

However, developing a framework to assure LLM-enabled CPS is challenging due to two key challenges. 
The first challenge is to ensure that LLMs do not violate critical specifications, such as safety and temporal constraints.
Due to the aforementioned phenomenon of hallucination, LLMs are inherently susceptible to producing incorrect or misleading outputs that may violate these specifications.
Moreover, the vast number of parameters in LLMs makes direct verification of the models practically infeasible.
Compounding the difficulty, LLM-enabled CPS often involve natural language as the inputs and outputs, which fall outside the scope of traditional verification tools that are not equipped to assess compliance with specifications and outputs expressed in natural language.
The second challenge stems from the inherent limitations of LLM capabilities.
Despite substantial advancements, LLMs still struggle with complex reasoning tasks, such as those requiring multiple goals and constraints satisfaction, which undermines their effectiveness in generating correct solutions for complex planning problems.
Several techniques, such as retrieval augmented generation (RAG)~\cite{xu2024p} and chain-of-thought prompting (CoT)~\cite{stechly2024chain}, have been proposed to partially mitigate these limitations in certain scenarios.
However, they only address the problem at a superficial level.
As task complexity continually increases, such techniques become inadequate, often failing to produce reliable or correct outcomes.

To address the above challenges, we propose our hierarchical neuro-symbolic framework, SafePilot, which aims to provide comprehensive and general assurance for LLM-enabled CPS.
Our framework takes as input a task description and corresponding specifications, such as safety and temporal specifications, expressed in natural language under a predefined format, and generates a plan that satisfies the given specification.
Our system is composed of several key components.
The first is \textbf{hierarchical planner}, which operates at a high level of abstraction.
It does not generate detailed plans directly.
Instead, it is responsible for decomposing and composing tasks and plans. 
It receives the initial task description and specifications from the user and first passes them through a discriminator to assess the difficulty of the task for the LLM.
If the task is determined to be too challenging, the hierarchical planner is activated to address the problem using a divide-and-conquer strategy. 
Specifically, it decomposes the original task into a series of sub-tasks, which are individually solved by other components.
The resulting sub-plans are then composed into a final plan by the hierarchical planner.
The second is \textbf{verifier}.
It first receives the specifications expressed in natural language and leverages the LLM to translate them into formal formulas.
The verifier is then responsible for checking whether a given plan satisfies these formal formulas, that is, whether the plan adheres to the specifications.
The third is \textbf{task planner}.
It receives a task description from the hierarchical planner and uses the LLM to generate a step-by-step plan candidate.
This candidate is passed to the verifier for validation.
If the plan fails to meet the specifications, the system constructs a new prompt based on the verification feedback to guide the LLM in generating a revised plan candidate. 
This iterative process continues until a plan satisfying the specifications is produced or a predefined iteration limit is reached.
All sub-plans and the final plan are subjected to verification by the verifier to ensure strict adherence to the given specifications.
Through this approach, our hierarchical neuro-symbolic framework, SafePilot, is capable of solving complex planning problems while ensuring strict compliance with given specifications.

In summary, our research makes several key contributions:
(1) We propose SafePilot, a novel hierarchical neuro-symbolic framework that provides end-to-end assurance for LLM-enabled CPS.
(2) We address the critical assurance problem in LLM-enabled CPS by introducing a verification-guided planning mechanism.
(3) We enhance the capability of LLM-enabled CPS in solving complex tasks through a hierarchical planning strategy.
(4) We demonstrate that SafePilot is both general and effective through extensive case studies involving complex planning tasks.

The remainder of the paper is organized as follows. 
Section~\ref{related} presents the related works.
Section~\ref{preliminaries} gives the preliminaries of this paper.
Section~\ref{system} provides the details of system design.
Section~\ref{experiments} evaluates our framework using two case studies.
Section~\ref{conclusion} concludes the paper.

\section{Related Works}~\label{related}
In this section, we first introduce applications of LLM-enabled CPS, then discuss certain LLM-enabled CPS where the LLM is responsible for reasoning and planning, and finally review works on verifying certain LLM-enabled CPS, comparing it with our own approach.

\textbf{LLM-enabled CPS.}
LLM-enabled CPS are applied across various fields, and the role of LLMs in these systems varies significantly.
In some LLM-enabled CPS, LLMs assist with tasks such as data processing and context grounding. 
They do not make specific decisions within the system but provide support to the system.
However, this paper primarily focuses on applications where LLMs serve as the "brain" of the system~\cite{xu2024llm}, meaning that these systems leverage the reasoning and planning capabilities of LLMs. 

Researchers have leveraged LLMs in systems for reasoning and planning.
Traditional planning methods~\cite{antonyshyn2023multiple, erol1995hierarchical, yang2016survey, tang2012review} require human experts to carefully design solutions for specific problems and implement algorithms using programming languages, such as C++ or Java.
In contrast, LLMs possess versatility and ease of use, allowing them to generate plans for different problems by constructing prompts in natural language tailored to specific scenarios.
Jansen~\cite{jansen2020visually} demonstrates the capability of LLMs to generate high-level instructions for robots based solely on natural language inputs.
Dilu ~\cite{wen2023dilu} presents the concept of integrating LLMs as decision-makers in autonomous vehicles, enabling them to generate sequences of actions.
RT-2~\cite{brohan2023rt} develops a multimodal LLM capable of processing images and user instructions as inputs to generate plans in an end-to-end approach. 
Similarly, PaLM-E~\cite{driess2023palm} introduces embodied language models that integrate continuous real-world sensor data directly into the LLM, enhancing its ability to perceive and interpret environmental contexts.
Despite these advancements, the reliability of LLM-enabled CPS has consistently been a concern.
Due to the interaction between CPS and the physical world, reliability is a crucial aspect of their design and operation.
Without guarantees of reliability, a CPS may enter an unsafe state or miss deadlines, potentially leading to significant losses.
Therefore, verifying LLM-enabled CPS is a necessary prerequisite before they can be deployed in real-world applications.
However, due to the vast number of parameters in LLMs, verifying an LLM directly is clearly an impractical task.

\textbf{Assuring LLM-enabled CPS.}
Some researchers are focusing on assuring LLM-enabled CPS.
Ahn et al.~\cite{ahn2022can} propose a method to leverage LLMs for generating plans in robotic applications. 
At each timestep, they evaluate the usefulness and feasibility of all potential skills suggested by the LLM, executing the one with the highest score. 
However, their approach focuses only on the validity of each individual step, whereas systems, such as robotics, typically require a plan consisting of multiple steps. 
Their method cannot ensure that the final plan is both useful and feasible.
Lin et al.~\cite{lin2023text2motion} extend this approach by incorporating multi-step planning. 
Despite that, these methods do not provide formal guarantees, and their focus is limited to attribute-based constraints.

Some researchers go further, using formal logic to assist in verifying LLM-enabled CPS.
Jha, et al.~\cite{jha2023neuro} propose a counterexample guided synthesis framework for motion planning.
Their method provides formal guarantees, while limited to attribute-based constraints expressible through first-order logic (FOL)~\cite{barwise1977introduction}.
Their approach merely uses counterexamples to guide the LLM toward better outcomes, rather than providing detailed reasoning, this results in only limited improvements in the LLM with each iteration.
Yang, et al.~\cite{yang2024plug} propose a framework to reduce the ambiguity to enforce the temporal constraints for LLM-driven robot agents.
Even so, their approach has notable issues. 
First, their approach relies solely on Linear Temporal Logic (LTL)~\cite{bauer2010comparing}, a type of temporal logic that can effectively express constraints such as deadlines or access sequences. 
However, many attribute-based constraints, such as the distance to obstacles, cannot be well-expressed in LTL.
These constraints are common safety constraints in systems, such as robotics or autonomous driving, and can be expressed by FOL.
Second, their method verifies each step of the plan as it is generated.
If a step fails verification, the state reached by the previously verified steps is treated as the initial state.
Then the LLM is queried to generate new plan steps from this initial state, with verification continuing until the target state is reached.
This approach gradually progresses toward the target state through verified steps.
However, it has significant drawbacks.
For example, the verified steps generated by the LLM might guide the system to an initial state from which the target state cannot be reached, leading to the failure of the entire plan.

Existing methods~\cite{ahn2022can, lin2023text2motion, jha2023neuro, yang2024plug} struggle to fully express attribute-based and temporal constraints in CPS. 
Our approach, however, supports both FOL and LTL, effectively capturing these constraints.
Some methods verify individual steps, which may cause the overall plan to violate constraints~\cite{ahn2022can} or make it challenging to generate a complete plan~\cite{yang2024plug}. 
Our method, on the other hand, considers the plan as a whole, prompting the LLM to generate the entire plan from scratch each time, thereby avoiding these issues.
Compared to approaches~\cite{jha2023neuro, yang2024plug} that use only counterexamples from failed verifications to iteratively prompt the LLM, our method goes a step further by generating natural language-based reasoning from the verification process. 
This reasoning includes not only steps that adhere to constraints throughout the plan but also highlights the steps that violate constraints and explains the reasons for the violations. 
This approach effectively enhances the LLM’s success rate.

\textbf{LLMs in Complex CPS Tasks.}
To address more complex problems, some approaches employ techniques such as Chain-of-Thought (CoT)~\cite{stechly2024chain} prompting or Retrieval-Augmented Generation (RAG)~\cite{xu2024p} to enhance the reasoning capabilities of LLMs, with the aim of producing more accurate and reliable results.
Some approaches, such as~\cite{wang2023conformal}, adopt a step-by-step execution of sub-tasks to handle a complete task.
However, the sub-tasks are manually decomposed by the user before sending to the system, and the relationships among sub-tasks are organized using user-defined LTL formulas. 
This reliance on manual specification raises the barrier to use and limits the method’s general applicability. 
Furthermore, their approach can only provide probabilistic guarantees.
In contrast, our approach does not require manual decomposition of sub-tasks by the user and is capable of providing assurance of specifications, thereby improving both usability and reliability.

In summary, existing approaches fall short in two key aspects: they are unable to simultaneously and effectively handle both safety (attribute-based) and temporal specifications, and they lack the ability to scale to complex planning problems. 
Our proposed framework addresses both challenges by supporting specifications expressed in both FOL and LTL, ensuring comprehensive compliance, and by introducing a hierarchical planning strategy that enables the system to solve complex tasks through automated decomposition and iterative refinement. 
This allows our method to provide strong reliability guarantees while maintaining scalability and usability across diverse CPS.

\section{Preliminaries}~\label{preliminaries}
In this section, we first introduce the two types of formal logic used in this work and then present the problem statement.

\textbf{First-order Logic.}
First-order logic (FOL) is a formal system that extends propositional logic with the ability to quantify over individual elements within a domain. A typical first-order language consists of:
A set of \emph{terms} representing objects in the domain, which can be variables, constants, or function symbols applied to other terms.
A set of \emph{predicate symbols} that can be used to express properties of, or relations among, the objects denoted by the terms.
Logical connectives (e.g., $\land$, $\lor$, $\neg$, $\rightarrow$) and quantifiers (i.e., $\forall$ for ``for all'' and $\exists$ for ``there exists'').

A formula in first-order logic can assert that certain predicates hold (or do not hold) for specific combinations of terms. 
By using quantifiers, one can express statements about all objects in the domain or the existence of some objects satisfying particular properties. 
This expressive power makes FOL especially useful for specifying constraints in planning or synthesis problems, where the ability to quantify over objects and relations is often crucial.
In our work, FOL is used to formally describe some safety-related constraints.
Examples include avoiding collisions with obstacles, or ensuring that the robotic arm is not already holding a block before attempting to pick one up.

\textbf{Linear Temporal Logic.}
Linear temporal logic (LTL) is a modal temporal logic designed to reason about the behavior of systems over time. In contrast to purely FOL, LTL formulas incorporate temporal operators that capture how properties evolve along a linear timeline (i.e., a single sequence of states). The key temporal operators include: $\mathbf{X}\varphi$ (Next): $\varphi$ must hold in the next state.
$\mathbf{G}\varphi$ (Globally): $\varphi$ must hold in all future states.
$\mathbf{F}\varphi$ (Eventually): $\varphi$ must hold at some state in the future.
$\varphi \mathbf{U} \psi$ (Until): $\varphi$ must hold continuously until $\psi$ eventually becomes true.

LTL is widely used in the formal specification and verification of reactive systems, where it is crucial to ensure that certain temporal requirements hold throughout an execution. 
By leveraging LTL, one can succinctly capture constraints that must persist or eventually occur, making LTL a natural fit for modeling and verifying temporal properties in planning tasks.
In our work, LTL is used to formally describe the temporal-related constraints.
Examples include visiting city A before visiting city D, or not visiting city B in your travel.

\textbf{Problem Statement.}
We begin by introducing the key concepts and notations motivated by the fields of formal synthesis and planning.
Formally, we define our planning problem as a tuple $\langle \mathcal{S}, \mathcal{A}, \mathcal{T}, s_0, \mathcal{S}^g, \mathcal{C} \rangle$
, where $\mathcal{S}$ is a finite, discrete set of states capturing the possible configurations of both the controllable agent and the environment. Depending on the domain, this can include observable variables of the agent itself (e.g., positions, statuses) and any relevant aspects of the environment (e.g., obstacles, events). 
$\mathcal{A}$ is the set of symbolic actions that the controllable agent is able to perform. Each action $a \in \mathcal{A}$ generally has \emph{preconditions} (i.e., the requirements under which it can be executed) and \emph{effects} (i.e., how it modifies the state).
$\mathcal{T}: \mathcal{S} \times \mathcal{A} \rightarrow \mathcal{S}$ is the state transition function describing how the environment evolves when an action $a$ is applied in a state $s$. Specifically, if action $a$ is feasible in state $s$, then $\mathcal{T}(s, a)$ returns the resulting state.
$s_0 \in \mathcal{S}$ is the \emph{initial} world state from which planning begins.
$\mathcal{S}^g \subseteq \mathcal{S}$ is the set of \emph{goal states}. Each goal state must satisfy a list of user-defined goal conditions (e.g., propositions) combined by logical operators such as \texttt{and}, \texttt{or}, and \texttt{not}.
$\mathcal{C}$ represents additional \emph{specifications} that must hold throughout plan execution, encompassing both logical and temporal constraints. We may view $\mathcal{C}$ as defining a subset $\mathcal{S}^C \subseteq \mathcal{S}$ where every state in $\mathcal{S}^C$ satisfies these specifications. A valid plan must ensure that all visited states lie in $\mathcal{S}^C$.

A \emph{solution} to this planning problem is a symbolic plan
\begin{align*}
\small
&\pi = \langle a_1, a_2, \ldots, a_n \rangle,\quad a_i \in \mathcal{A} \quad \text{for all } i = 1, \ldots, n - 1. \\
s.t. \quad &s_i = \mathcal{T}(s_{i-1}, a_i) \quad \text{for all } i = 1, \ldots, n - 1, \\
&s_i \in \text{Pre}(a_{i+1}) \quad \text{for all } i = 0, \ldots, n - 1, \\
&s_i \in \mathcal{S}^C \quad \text{for all } i = 0, \ldots, n, \\
&s_n \in \mathcal{S}^g.
\end{align*}

The feasibility conditions for $\pi$ are as follows:
Throughout the execution of the actions $a_1, a_2, \ldots, a_n$, the system transitions though the corresponding states $s_1, s_2, \ldots, s_n$.
$\text{Pre}(\cdot)$ represents the preconditions of an action.
The preconditions of $a_{i+1}$ must be satisfied in the state resulting from applying $a_i$, for $i=0,\ldots,n-1$.
Every intermediate state $s_i$ must belong to $\mathcal{S}^C$, ensuring that all states satisfy the required specifications $\mathcal{C}$.
The final state, obtained after applying $a_n$, must lie in $\mathcal{S}^g$, thus meeting all goal conditions.

In other words, the plan must not only achieve the designated goals but also ensure that no specification constraints are violated at any stage. In this paper, we investigate methods to synthesize such a plan efficiently, respecting both the logical and temporal requirements encapsulated in $\mathcal{C}$.
\section{System Design}~\label{system}
This section begins with an overview of the entire framework. 
We then provide a detailed explanation of the LLM-driven task planning and verification process within our system. 
Finally, we present our hierarchical planning methods.

\subsection{Overview}
\begin{figure}[h]
    \centering   
    \includegraphics[width=\columnwidth]{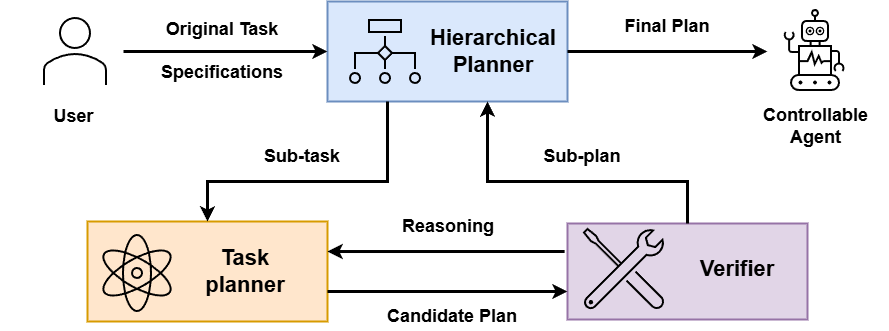}
    \caption{The overview of our hierarchical neuro-symbolic LLM-enabled planner.}
    \label{fig:framework}
\end{figure}
Fig.~\ref{fig:framework} illustrates the framework of our proposed hierarchical neuro-symbolic LLM-enabled planner, SafePilot, which consists of three main components: the hierarchical planner, the task planner, and the verifier, shown in blue, orange, and purple, respectively.
The SafePilot receives the task description and specifications in natural language from the user, and finally outputs a plan to the controllable agent.
The plan not only satisfies the task's goal but also follows the specifications, including both temporal and logical specifications.

It is worth noting that our method represents concrete tasks using a format similar to PDDL (Planning Domain Definition Language)~\cite{aeronautiques1998pddl}, a standardized language commonly used in automated planning. 
Such structured representations not only facilitate LLM comprehension but also enable accurate and automated processing by our verifier. 
While we adopt a PDDL-like format for clarity and consistency, the approach is flexible and can accommodate other structured representations depending on the implementation, such as JSON.
Prior works~\cite{silver2022pddl, gestrin2024nl2plan} have explored the integration of LLMs with PDDL by prompting the LLM to generate complete PDDL files from natural language descriptions, which are then processed by conventional PDDL-based planners. 
However, these approaches often suffer from the limited accuracy of LLM-generated PDDL, necessitating expert review and incurring significant manual effort.
In contrast, our approach requires users only to structure the problem in a specific format, while allowing other components—such as action descriptions—to remain in natural language. 
This design choice significantly lowers the barrier to use and enhances overall usability.

The \textbf{hierarchical planner} operates at an abstract level. 
Rather than generating detailed plans on its own, it orchestrates the entire planning process by decomposing and composing tasks. 
Upon receiving the initial task and specifications, the hierarchical planner first invokes a discriminator that gauges the complexity of the task for the LLM. 
If the task is deemed too challenging, the hierarchical planner steps in to manage the problem via a strategy akin to dynamic programming, subdividing the original task into multiple sub-tasks. 
Each sub-task is then handed off to other components for resolution, and their respective sub-plans are composed into an overall plan. 
Otherwise, the hierarchical planner delegates the original task directly to the other components for processing.
By initially evaluating the complexity of a given task, the system can determine whether decomposition is required, thereby preventing unnecessary time and computational overhead caused by LLMs failing to solve overly complex problems. 
This divide-and-conquer strategy not only enhances efficiency but also provides a systematic method for effectively harnessing LLMs to address challenging tasks.
Once the hierarchical planner sends a sub-task or the original one, the \textbf{task planner} generates a detailed, step-by-step plan candidate. 
This candidate is then passed to the verifier for validation. 
If it fails any specification, the verification feedback is used to construct a refined prompt for the LLM, prompting it to revise the plan. 
This iterative loop continues until a valid plan is found or a predefined iteration limit is reached. 

The \textbf{verifier} enforces compliance with the given temporal and logical specifications.
It begins by translating the user-provided natural language specifications into formal formulas using the LLM, ensuring that safety and temporal requirements are expressed precisely. 
Whenever a candidate plan is proposed from the task planner, the verifier checks it against these formal specifications using formal verification.
This check determines whether the plan is valid or violates any constraints.
If a violation is detected, a reasoning prompt is generated based on the verification process to guide the LLM in producing a revised plan.

Throughout this process, all sub-plans and the final composed plan must pass verification, ensuring they rigorously adhere to the user-defined requirements.
\subsection{LLM-driven Task Planning with Verifications}
In this subsection, we introduce how the task planner and the verifier operate.
They receive either decomposed sub-tasks or the original task from the hierarchical planner, and leverage the LLM along with the verification tool to generate a plan that satisfies the given specifications, which is then returned to the Hierarchical Planner.

\begin{figure}[h]
    \centering   
    \includegraphics[width=\columnwidth]{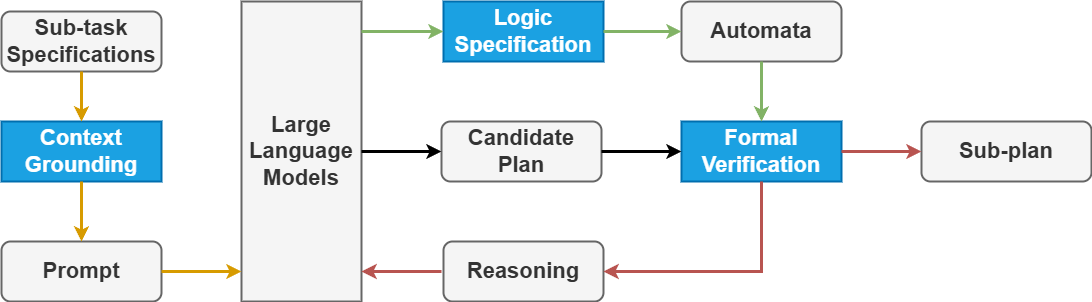}
    \caption{LLM-driven Task Planning with Verification. Here we use a sub-task received from hierarchical planner.}
    \label{fig:sub framework}
\end{figure}
Fig~\ref{fig:sub framework} illustrates the LLM-driven task planning with verification, 
comprising four primary processes highlighted in blue: context grounding, logic specification, and formal verification.
For illustration purposes, here we use a sub-task received from the hierarchical planner as an example.

The \textbf{context grounding} process, illustrated by yellow paths, is a shared procedure between the task planner and the verifier.
It takes the original task or the sub-task and the natural language specifications as the input, and translates to two well-designed prompts through prompt engineering~\cite{liu2023pre, marvin2023prompt} as the output.
The first prompt is to require the LLMs to translate natural language-based specifications $\mathcal{C}$ into formal specifications expressed as logical formulas. 
The generated logical formulas are then reviewed by domain experts, with the detailed process described later.
The second one is to instruct the LLM to generate multi-step plans that achieve the target and satisfy these specifications.

From a symbolic perspective, the second prompt represents a set of all possible plans.
If the prompt design is insufficient, preventing full grounding, may lead to fundamental errors, corresponding to lead the LLMs to produce incorrect results.
Typically, we begin with straightforward examples to help the LLM understand the context.
This is a widely used method for constructing effective prompts~\cite{ahn2022can, lin2023text2motion, jha2023neuro}.
Subsequently, we guide the LLM to imitate this process in accomplishing the actual tasks.
Here is an example of an initial prompt.
\begin{tcolorbox}[title=Example of an Initial Prompt, breakable, before title=\vspace{-1mm}, after title=\vspace{-1mm}, top=0mm, bottom = 0mm]\label{box1}
\textbf{\# Tasks background.} \\
You are a planner for drivers. 
There are several cities on the map and some paths between these cities, for example, A-B means there is one path between city A and city B.\\
...\\
\textbf{\# Permitted actions.} \\
The driver is able to take a path multiple times and can visit a city multiple times. \\
...\\
\textbf{\# Example problem.} \\
Here is an example problem and the correct result.\\
Given the planning problem driver-0\\
...\\
\textbf{\# Example result.} \\
The solution for the problem driver-0 is:\\
START-PLAN\\
1. A -> B\\
...\\
\textbf{\# Problem.} \\
Now please give me the result of the new planning problem driver-1 below, the solution's format should be the same as the example solution:\\
Given the planning problem driver-1\\
(define (problem driver-1)\\
...\\
\end{tcolorbox}
As shown in the example above, we first provide the task background, followed by the operations the LLM is permitted to use. 
Next, we present a sample problem along with the correct answer in the specified format.
Finally, we provide the problem to be solved, instructing the LLM to respond in the same format as the sample problem.


For the \textbf{logic specification} process, marked by green paths, it takes the logic formula $\boldsymbol{\varphi}$ output from the LLM as input, which after experts reviewing, is converted into an automaton (for temporal specifications) as the output.
Here we use the temporal specifications as an example.
The output for logical specifications, similar to that for temporal specifications, depends on the specific verification tool used, which will be explained in detail later.
This is a process operating in the verifier component.
Specifically, before the logic specification process runs, the LLM receives a prompt related to the specifications and outputs a corresponding logic formula, such as LTL. 

\begin{figure}[h]
    \centering   
    \includegraphics[width=\columnwidth]{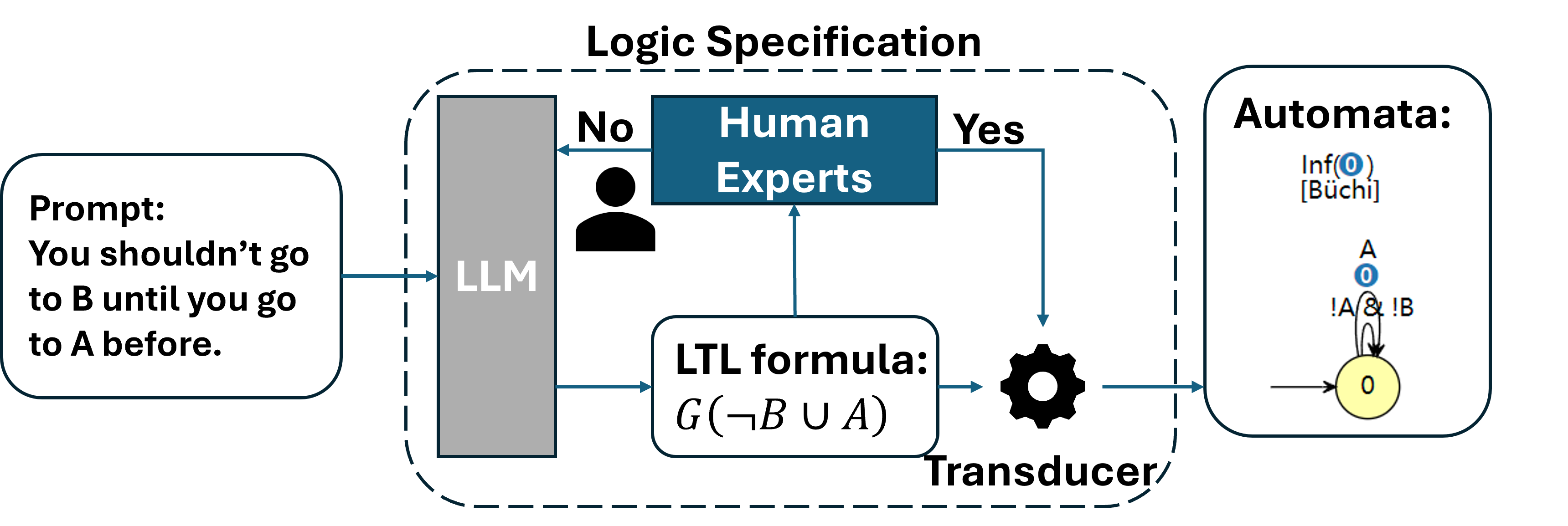}
    \caption{Illustration of Logic Specification. Here we use an LTL formula as an example of the logic formulas.}
    \label{fig:logic spec}
    \vspace{-1mm}
\end{figure}

Fig.~\ref{fig:logic spec} shows the process of the logic specification component, where $A$ and $B$ here indicate whether the driver has visited cities $A$ and $B$.
As you can see, the LLM takes natural language-based specifications $\mathcal{C}$ as input and outputs a logic formula $\boldsymbol{\varphi}$.
In the figure, we use an LTL formula for illustration.
This logic formula $\boldsymbol{\varphi}$ is then reviewed by a human expert.
If it does not pass the review, the LLM is prompted to regenerate it until it is approved.
After passing the human expert review, the logic formula $\boldsymbol{\varphi}$ is sent to a transducer, which automatically converts it into an automaton $A$.
This automaton $A$ is the final output of the logic specification process.


It is important to note that for the same tasks and constraints, the required automaton $A$ is identical, meaning the formal specification component only needs to execute once.
This consistency of the automaton across iterations minimizes manual effort.
This automaton represents a set of plans that satisfy the specifications.
Upon completion of context grounding and logic specification, the LLM generates a candidate plan for the controllable agent, though its compliance with constraints is yet to be confirmed.


The \textbf{formal verification} process, depicted in red paths, receives two inputs: the candidate plan and the automaton derived from the specifications. 
This component can be viewed as a function which outputs True or False according to the verification result.
It utilizes formal verification tools, such as Z3 Python API~\cite{z3prover} or Spot Python~\cite{duret.22.cav}, to ascertain whether the plan breaches the formal specifications.
Z3 is designed as a high-performance SMT (Satisfiability Modulo Theories) solver, which extends FOL with additional theories, making it highly versatile for various applications.
It is widely used in fields like formal verification, automated reasoning, and constraint solving.
Spot is an open-source library and tool designed for the manipulation, verification, and optimization of automata and temporal logic formulas.
It is popular in fields of formal verification, particularly for handling LTL and other temporal logics, and is useful for creating, transforming, and verifying automata representations of temporal logic specifications.
Both of these tools provide comprehensive documentation to help users quickly get started.
In SafePilot, Z3 Python API is employed for FOL verification, while Spot Python API is used for LTL verification. 
If the plan satisfies the formal verification, it is then sent back to the hierarchical planner component.
If it fails, the verification process yields detailed feedback, which is then transformed into a reasoning prompt.
Then the LLM could, based on its commonsense and reasoning capabilities, analyze the reasons, actively collect more information, and generate a more accurate and specification-compliant plan.

Here is a simple example from the case study of the navigation problem.
\begin{tcolorbox}[title=Reasoning Example, before title=\vspace{-1mm}, after title=\vspace{-1mm}, breakable, top=0mm, bottom = 0mm]\label{box1}
D $\rightarrow$ B \\
B $\rightarrow$ C \\
is invalid. \\
It violates the constraint: \\
You shouldn't go to B until you go to A before.
\end{tcolorbox}
In this reasoning prompt, we first identify which step in the plan is incorrect and then specify the corresponding constraint that has been violated.
This detailed reasoning greatly contributes to improving the output of the LLM.
This point will be demonstrated through experiments later.

The LLM continuously refines its output until it either passes verification or reaches the predefined iteration limit.
It is important to note that our LLM-driven task planning with verifications can only guarantee that the outputted plan meets the requirements.
However, it cannot guarantee that a plan will always be outputted.
If the iteration limit is exceeded, it will output a failure.

\subsection{Hierarchical Planning}
Hierarchical planning is a process that is carried out at the beginning by the hierarchical planner component.
It begins with a \textbf{discriminator} that evaluates whether a task is suitable for direct LLM-based planning. 
This prevents repeated failed attempts and unnecessary computation. 
If the task is judged too difficult, it is decomposed into sub-tasks; otherwise, it is passed directly to the LLM-driven task planner with verification. 
The resulting sub-plans are later recomposed into the final solution through a divide-and-conquer strategy.

To assess task difficulty, directly scoring natural-language queries is unreliable due to their variability. 
Instead, we evaluate complexity from the LLM’s own perspective. 
By examining the average token-level output probabilities~\cite{kim2023better}, we estimate the model’s confidence in generating the intended answer. 
These probabilities are aggregated into a confidence score and compared with a user-defined threshold to determine whether decomposition is required.

For tasks requiring decomposition, the \textbf{decomposition and composition} module parses the goal into a Boolean logic tree (e.g., using AND/OR/NOT connectors). 
Each sub-goal becomes an independent sub-task, while shared components such as actions remain unchanged. 
Designing this process is non-trivial: (1) the task planner cannot guarantee a feasible plan for every sub-task, and (2) it may return different plans across iterations, altering the initial state for subsequent sub-tasks.
Thus, we incorporate a failure-resilient mechanism that supports backtracking.
When planning for a sub-task fails, the system revisits and revises previous sub-plans to restore a new valid initial state for the sub-task.
A user-defined retry limit determines how many times this process may repeat.
However, if the composed sub-plan ultimately fails to pass verification, the sub-goals and their corresponding sub-plans can serve as concrete in-domain examples to guide the LLM in solving the original goal directly. 
This plays a role similar to RAG, while avoiding the need to prepare large datasets or perform similarity-based retrieval.

For simplicity, Algorithm~\ref{algo} illustrates the procedure for goals connected by the AND operator.
Line 1–3 define the inputs and a global failure counter $f$ indexed by initial and goal nodes.
Line 4 performs AND-decomposition of the goal into a list of sub-goals $\mathcal{G}^{sub}$.
Line 5-6 initialize the current state, the accumulated plan, the sub-goal index, and the stack (for rollback)
The main loop (Lines 7–21) processes sub-goals sequentially.
Lines 9-13 handle the case where a sub-goal $G_k$ has failed too many times: the algorithm rolls back by popping from the stack and restoring a previous state and partial plan. 
If the stack is empty, the whole procedure fails.
Line 14 recursively invokes the hierarchical planner on this sub-goal.
Line 15-17 treat unsuccessful sub-task planning by incrementing the failure counter.
Line 18-21 handle successful sub-task planning.
The start point of the current sub-task is pushed onto the $\textsc{Stack}$ to facilitate potential revision later.
The sub-plan is concatenated into $\pi_{\text{total}}$, the failure counter for $(s, G_k)$ is reset, and the state is updated.
The algorithm then proceeds to the next sub-goal.

In line 4, the AND-decomposition may produce sub-goals that either have or do not have logical or sequential dependencies. 
When no dependencies exist, the sub-goals are solved independently, their sub-plans are concatenated, and the resulting plan is verified. 
If this verification fails, the generated sub-tasks and sub-plans serve as concrete examples to guide the LLM-based planner with verification to produce a corrected plan. 
When dependencies do exist, meaning that executing a later sub-task may affect the validity of an earlier one, the sub-plans cannot be composed directly. 
In this case, the sub-tasks and sub-plans again function as guiding examples for the LLM-based planner to synthesize a new valid overall plan.

\begin{algorithm}[t]
\caption{Hierarchical Task Decomposition and Composition}~\label{algo}
\begin{algorithmic}[1]
\State \textbf{Input}: Goal tree $G$, initial state $s_0$, max attempts $t$
\State \textbf{Output}: Plan $\pi$ and final state $s$, or $\emptyset$
\State \textbf{Global}: Failure counter $f$ (keyed by initial and goal nodes)

\State $\mathcal{G}^{sub} \gets \textsc{GetSubgoals}(G)$ \Comment{AND-decomposition}
\State $s \gets s_0,\ \pi_{\text{total}} \gets \emptyset,\ k \gets 0$ \Comment{idx, state, plan}
\State $\textsc{Stack} \gets \emptyset$ \Comment{Stores $(k, s, \pi)$}


\While{$0 \le k < |\mathcal{G}^{sub}|$}
    \State $G_k \gets \mathcal{G}^{sub}[k]$ \Comment{Process a sub-task}

    \If{$f[(s, G_k)] \ge t$}  \Comment{Too many failures on $(s, G_k)$}
        \If{$\textsc{Stack} = \emptyset$}
            \State \Return $(\pi_{\text{total}}, \emptyset)$ \Comment{No more rollback}
        \EndIf
        \State $(k, s, \pi_{\text{total}}) \gets \textsc{Stack.pop}()$ \Comment{Rollback}
        \State \textbf{continue}
    \EndIf

    \State $(\pi_{\text{sub}}, s') \gets \textsc{HierarchicalPlanner}(G_k, s, t)$

    \If{$s' = \emptyset$} \Comment{Sub-task failed}
        \State $f[(s, G_k)] \gets f[(s, G_k)] + 1$
        \State \textbf{continue}
    \Else \Comment{Sub-task success}
        \State $\textsc{Stack.push}((k, s, \pi_{\text{total}}))$ \Comment{Record start point}
        \State $\pi_{\text{total}} \gets \pi_{\text{total}} \oplus \pi_{\text{sub}}$
        \State $f[(s, G_k)] \gets 0$, $s \gets s'$, $k \gets k + 1$ \Comment{Reset counter}
    \EndIf
\EndWhile

\State \Return $(\pi_{\text{total}}, s)$
\end{algorithmic}
\end{algorithm}

Through the hierarchical planning, we provide a feasible solution for enabling LLMs to tackle complex planning problems.
The final composed plan adheres to both the specifications and the final goal, thereby ensuring the overall safety of the LLM-enabled CPS.
\section{Experiments}~\label{experiments}
In this section, we use two case studies to illustrate the effectiveness of our framework.
The first case involves a Blocksworld problem using FOL, which exemplifies issues where constraints can be expressed using FOL, such as a robot navigating a maze or avoiding obstacles. 
The second case employs LTL for a navigation problem, representing issues where constraints can be expressed using LTL, such as robotic motion planning with deadlines or event order constraints.
Additionally, FOL is employed in this case to handle certain logical specifications.

In the experimental process, our framework is already equipped to handle these issues.
We only need to construct the initial prompt based on the problem, and the framework can operate automatically to yield the final result.
We conduct experiments using two LLMs, GPT-4o and GPT-3.5-turbo.
It is important to note that our method can be easily extended to other LLM-enabled CPS with different LLMs and tasks.
Due to space limitations in the main text, we present only a portion of the prompts.

\subsection{Blocksworld Problem}
\subsubsection{Background}
In this case study, we address the Blocksworld~\cite{gupta1992complexity} problem, which is a classic planning problem for robotic arms, with primary constraints based on various attributes, such as action preconditions.
This problem involves a set of blocks, typically positioned on a flat surface, and the goal is to rearrange these blocks from an initial state to a target state using a specified set of robotic arm actions, including pick up, put down, stack, and unstack.
Each operation of the robotic arm is governed by corresponding logical constraints, which can be expressed using FOL.

\begin{figure}[h]
    \centering   
    \includegraphics[width=\columnwidth]{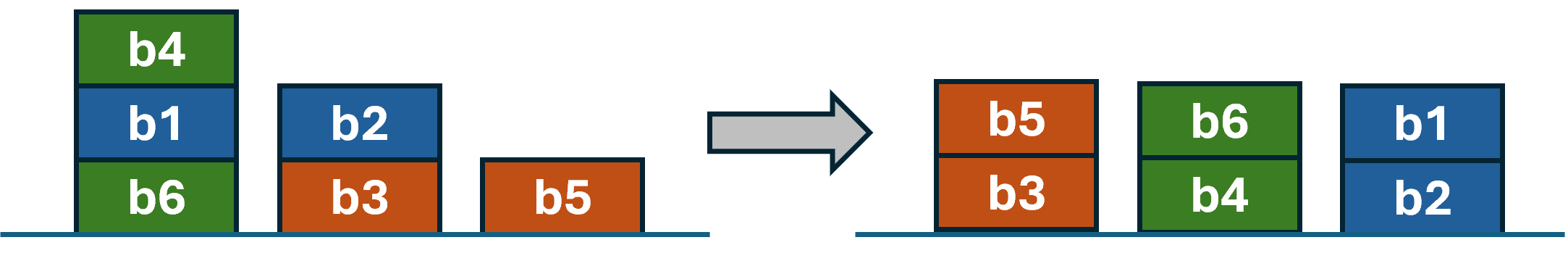}
    \caption{An example of a Blocksworld problem. The left side indicates the initial state, while the right side represents the goal state.}
    \label{fig:block}
\end{figure}

\subsubsection{Motivating Example} 
We provide an example to demonstrate the process and principles.
The initial and goal state of this problem are shown in Fig.~\ref{fig:block}.
If decomposition is required, it can be performed by first examining whether the statements connected by the AND operator in the goal have any logical dependencies. For example, in this case, the goal can be divided into three sub-goals, each corresponding to a separate stack of blocks.
However, treating these three statements as separate sub-goals may lead to multiple possible outcomes, as the target state doesn't clearly specify which blocks should be on the table. This ambiguity can cause the composed goal to fail verification. In such cases, the sub-goals and their sub-plans can serve as concrete examples to guide the LLM in solving the original goal directly.
For clarity of presentation, we use a relatively simple example here that does not require decomposition via the hierarchical planner.

Therefore, we directly use the LLM-driven planner with verification to deal with the example.
Initially, we establish the background knowledge and request the LLM to generate the relevant FOL constraints, expressed using the Z3 Python API.
This information is then saved into a file as the 1\st prompt.
For illustration purposes, only part of the prompt is shown in the experiment section.
The LLM generates the corresponding FOL based on the first prompt, expressed using the Z3 Python API, and it is confirmed as correct by a human expert in the formal specification component.
Subsequently, we construct the 2\nd prompt and store it in a file, shown below.
Similar to the 1\st prompt, this 2\nd prompt begins by introducing the background knowledge and permitted actions.
Then it provides an example problem along with the correct answer.
Finally, it presents the target problem and instructs the LLM to output the result in the same format as the example.

The framework takes these two prompt files and initiates its execution.
The framework initially outputs the FOL expressed in the form of the Z3 Python API.
After verification by the user, the LLM begins generating the plan, resulting in the 2\nd output, as shown on the left.
The framework uses the previously generated FOL to verify the produced plan. 
It can be found that the result violates the specifications, generates reasoning, and provides this as the 3\rd prompt.
In the 2\nd output, the plan attempts to stack block 5 on block 3 while block 2 is already on block 3, which violates the specification.
The LLM generates the 3\rd output, which, after verification, meets all specifications and is output as the final correct result to the user or agents.
This experiment illustrates that our approach can successfully extract the correct plan from an LLM while ensuring compliance with FOL constraints.

\begin{tcolorbox}[title=1\st Prompt, breakable, before title=\vspace{-1mm}, after title=\vspace{-1mm}, top=0mm, bottom = 0mm]\label{box1}
You are a motion planner who wants to use Z3 to verify your plan. Now we consider the Blocksworld planning problem. \\
The objects in the problem domain include a finite number of cubical blocks, and a table large enough to hold all of them. Each block is on a single other object (either another block or the table). For each block b, either b is clear or else there is a unique block a sitting on b. \\
...\\
Pick-up:
The block must be clear and on the table in state s1. The block must not be in the hand in state s1. The block must be in the hand in state s2. The block must not be on the table in state s2 and must not be clear.
For all other blocks x, the state of the hand, table, and clear should remain the same between s1 and s2.
For all blocks x, y, the state of the stacked should remain the same between s1 and s2.\\
...\\
Above are the details of these operations.
I will use Z3 to verify future plans. please help me complete the following code in which translate all the above four actions into first order logic using the  the formatting of Z3 Pyhton API.
Please do not alter the code within the provided class State.
I don't need any explanation.
Your response should include all the code, starting with 'from z3 import *'.
Below is the incomplete code. I have already implemented the pick\_up, stack functions for your reference...

% ...


% ...
\end{tcolorbox}

\begin{tcbitemize}[
  raster columns=2,
  raster equal height=rows,
  enhanced,
  raster column skip=1mm,
  top=0mm,
  bottom=0mm
]

\tcbitem[
  title={2nd Prompt},
  valign=top,
  before title=\vspace{-1mm},
  after title=\vspace{-1mm},
  halign=flush left
]
\ttfamily\small
...\\
(define (problem BW-rand-6) \\
(:domain blocksworld-4ops) \\
(:objects b1 b2 b3 b4 b5 b6) \\
(:init \\
(arm-empty) \\
(on b1 b6) \\
(on b2 b3) \\
(on-table b3) \\
(on b4 b1) \\
(on-table b5) \\
(on-table b6) \\
(clear b2) \\
(clear b4) \\
(clear b5) \\
) \\
(:goal \\
(and \\
(on b1 b2) \\
(on b5 b3) \\
(on b6 b4))))

\tcbitem[raster multicolumn=1, raster multirow=3, valign=top, blankest]
\begin{tcbitemize}[raster columns=1, raster equal height=rows, top=0mm, bottom=0mm]

\tcbitem[
  title={2nd Output},
  before title=\vspace{-1mm},
  after title=\vspace{-1mm},
  halign=flush left
]
\ttfamily\small
1. pick-up b5 \\
2. stack b5 b3 \\
3. unstack b1 b6 \\
4. pick-up b6 \\
5. stack b6 b4 \\
6. unstack b2 b3 \\
7. stack b2 b1

\tcbitem[
  title={3rd Prompt},
  before title=\vspace{-1mm},
  after title=\vspace{-1mm},
  halign=flush left
]
\ttfamily\small
1. pick-up b5 \\
2. stack b5 b3 \\
is invalid. \\
Step 2 violates the precondition of the action stack.

\tcbitem[
  title={3rd Output},
  before title=\vspace{-1mm},
  after title=\vspace{-1mm},
  halign=flush left
]
\ttfamily\small
1. unstack b4 b1 \\
2. put-down b4 \\
3. unstack b1 b6 \\
4. put-down b1 \\
5. pick-up b6 \\
6. stack b6 b4 \\
7. unstack b2 b3 \\
8. put-down b2 \\
9. pick-up b5 \\
10. stack b5 b3 \\
11. pick-up b1 \\
12. stack b1 b2

\end{tcbitemize}
\end{tcbitemize}

It is also possible to deploy SafePilot on a real testbed.
We demonstrate this feature using a real 6 DOFs (Degree of Freedoms) robotic arm to perform the same Blocksworld problem in Fig.~\ref{fig:block}.
The robotic arm is embedded with a Raspberry Pi and can execute corresponding actions based on the plan generated by the LLM for the Blocksworld problem.
We input the initial state of the blocks and the target state into SafePilot. 
The SafePilot generates a verified plan and outputs it to the robotic arm. 
The robotic arm then executes the plan step by step, adhering to the specified constraints and achieving the target state for the blocks.
\begin{figure}[h]
    \centering   
    \includegraphics[width=\columnwidth]{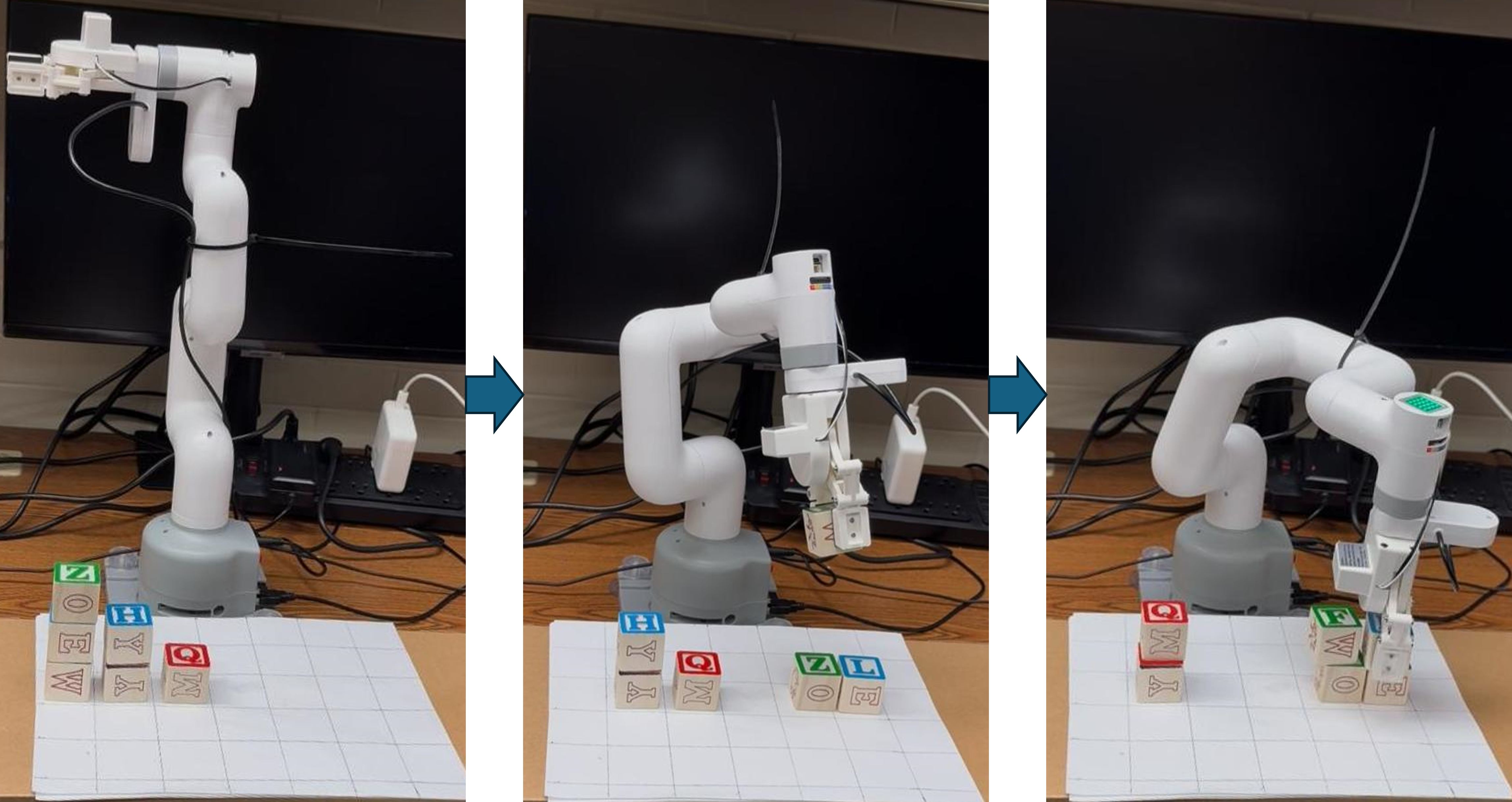}
    \caption{Deploy SafePilot on a robotic arm to solve a Blocksworld problem. The left image shows the initial state, the middle image shows the robotic arm executing the plan, and the right image shows the final state.}
    \label{fig:arm}
\end{figure}

\subsubsection{Quantitative Experiments}
We analyze the performance of our framework and the baselines on 50 random Blocksworld problems with different block numbers and different LLMs.
We select three types of baselines.
The first is a one-shot approach, which cancels the iterations in our LLM-driven task planner and directly verifies the LLM's first generated plan as the final result.
The second is the LLM-driven task planner in our proposed framework but without reasoning, which removes the reasoning process after verification, providing only the generated plan is invalid as the prompt for the next iteration with the LLM.
This baseline is similar to the method in~\cite{jha2023neuro}. Since this method~\cite{jha2023neuro} only supports FOL, it is suitable for the Blocksworld case study.
The third is the LLM-driven planner with verification in our proposed framework, which removes the hierarchical planner from SafePilot.
Our experiment uses two LLMs: GPT-4o and GPT-3.5-turbo.
In our selected problems, the number of blocks is 5, 7, 9, 11, and 13, with ten problems for each quantity.
The LLM-driven planner's iteration limitation has been set to 20.
The limitation $t$ of the number of failures for a single sub-task is set to 3.

\begin{figure}[htbp]
\centering
\begin{subfigure}{\columnwidth}
  \centering
  \includegraphics[width=\columnwidth]{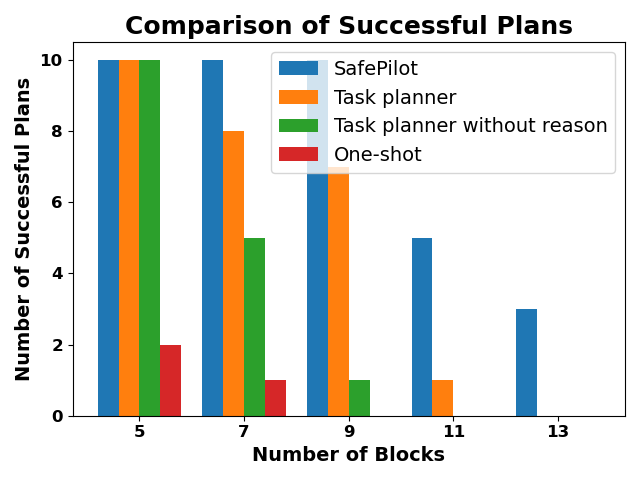}
  \caption{GPT-4o.}
  \label{fig:image1}
\end{subfigure}
\hfill
\begin{subfigure}{0.45\textwidth}
  \centering
  \includegraphics[width=\columnwidth]{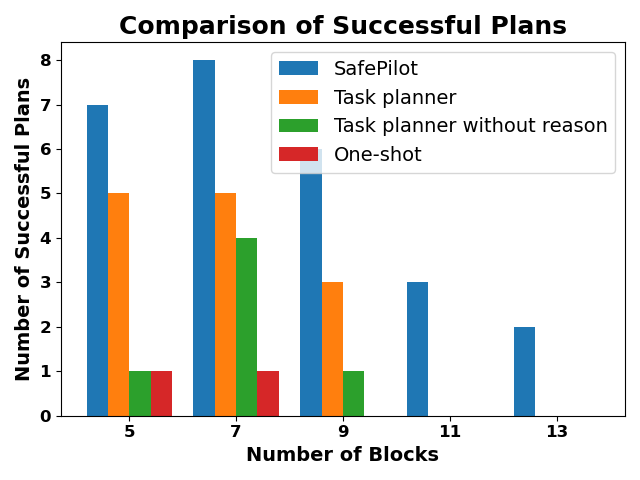}
  \caption{GPT-3.5-turbo.}
  \label{fig:image2}
\end{subfigure}
\caption{The performances of LLMs on Blocksworld problem. Fig.~\ref{fig:image1} uses GPT-4o, while Fig.~\ref{fig:image2} uses GPT-3.5-turbo. The blue bars represent the SafePilot, and the orange bars represent the task planner which removes the hierarchical planner from SafePilot. The green bars represent the task planner without reason, which is similar to method~\cite{jha2023neuro}. The red bars represent the one-shot method.}
\label{fig:blocks_exp}
\end{figure}

The results of quantitative experiments are shown in Fig.~\ref{fig:blocks_exp}.
First, the number of successful plans generated by SafePilot exceeds that of all other methods, especially as the number of blocks increases, which indicates a higher level of task complexity.
In such cases, other methods generally fail to produce correct plans, whereas SafePilot, through the use of the hierarchical planner, equips the LLM with the capability to effectively handle complex problems.
As the number of blocks increases, the number of successful plans generated by our SafePilot also shows a noticeable decline. 
This is because our hierarchical planner performs decomposition based on the goal, without altering the problem's world space.
As a result, although the sub-task goals are significantly simplified, the expanding world space reduces the likelihood of the LLM generating correct plans.
Second, the number of successful plans using task planner is noticeably higher than that of task planner without reasoning.
This demonstrates the effectiveness of using reasoning to guide the LLMs in updating the plan.
Third, compared to other methods, the success rate of the one-shot approach is very low.
It is nearly impossible to generate a valid plan as the number of blocks increases.
This demonstrates that iteratively prompting the LLM is very helpful in improving results.
Given the probabilistic nature of LLMs, only one attempt may fail to yield correct results.
This also demonstrates that the chosen Blockworlds problem is not trivial for the LLM.
In addition, we can also observe that the overall number of successful plans of GPT-4o is significantly higher than that of GPT-3.5-turbo, which is expected, as GPT-4o has more parameters and greater reasoning capability.

\begin{figure}[h]
    \centering   
    \includegraphics[width=\columnwidth]{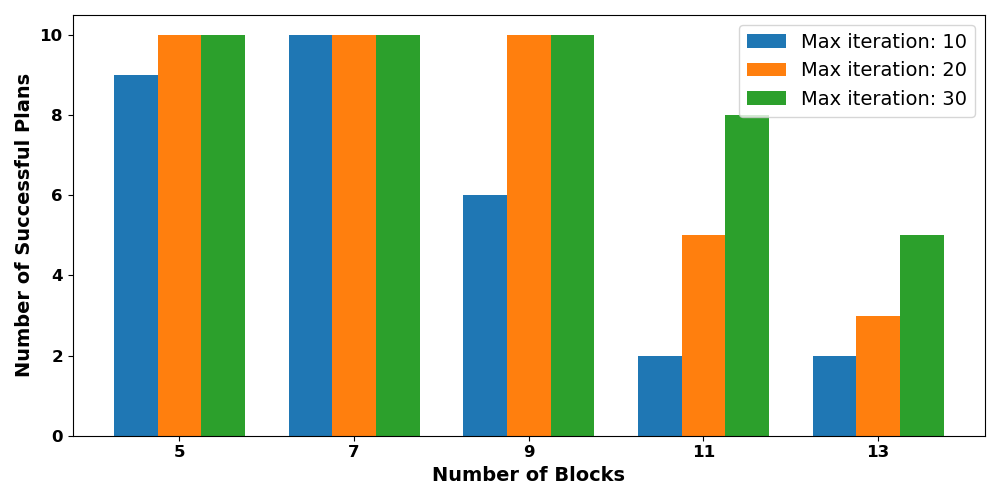}
    \caption{The performance of GPT-4o on Blocksworld problem using SafePilot with different iteration limits. The blue, orange, green, and red bars represent iteration limits of 10, 20, and 30, respectively.}
    \label{fig:block_limit}
\end{figure}

We also perform experiments to analyze the impact of iteration limits on the number of successful plans generated by the LLM.
We used SafePilot with GPT-4o to test the results for 50 Blocksworld problems with iteration numbers set to 10, 20, and 30. 
Among the 50 problems, the number of blocks was 5, 7, 9, 11, and 13, with ten problems for each quantity.
The results are shown in Fig.~\ref{fig:block_limit}.
From the figure, we can observe that higher iteration limits lead to a greater number of successful plans.
This effect is especially pronounced as the problems become more complex, i.e., as the number of blocks increases.
This is because more complex problems require stronger planning capabilities, and higher iteration limits enable SafePilot to exhibit enhanced planning capability.

\subsection{Navigation Problem}
\subsubsection{Background}
In this case study, we demonstrate the applicability of our approach in handling planning problems that involve timing-related constraints. 
We address a navigation problem that requires the LLM to devise a plan for a driver while adhering to temporal constraints, which can be expressed by LTL.
This is a classic navigation-related planning problem, with constraints that include both attribute-based factors, such as the presence of pathways, and time-related constraints, such as the sequence of visits. 
This problem involves several cities, some of which are connected by roads. 
The driver aims to reach several target cities.
The driver cannot traverse between cities without a direct path in a single step. 
We task the LLM with producing a plan for the driver to reach the target cities without violating the requirements. 

\subsubsection{Motivating Example} 
We present a detailed example to illustrate the process and underlying principles.
Similar to the Blocksworld problem, we use a relatively simple example here that does not require decomposition via the hierarchical planner for clarity of presentation.
We initially introduce the problem context and request the LLM to generate logic formulas representing the constraints.
The temporal requirement in this illustrative example is "You should have been to C and D before you go to G".
This requirement is asked to be specified as an LTL formula.
It will then be translated to an automaton expressed using the Spot Python API for further verification.
The other attribute-based requirements, such as the precondition for traveling from city A to city B, are specified as FOL constraints and expressed using the Z3 Python API, similar to the Blocksworld problem.
This content is saved as the 1\st prompt.
Subsequently, we provide an example with the correct outcome and instruct the LLM to solve a new problem following the same format.
The content of the second prompt is also preserved in a file, as shown in the 2\nd prompt below.
The map of this problem is shown as Maps below.

\begin{tcolorbox}[title=2\nd Prompt, breakable, before title=\vspace{-1mm}, after title=\vspace{-1mm}, top=0mm, bottom = 0mm]\label{box1}
You are a planner for drivers. 
There are several cities on the map and some paths between these cities, for example, (road A B) means there is one path between city A and city B.
All the paths will be provided to you.
The driver is able to take a path multiple times and can visit a city multiple times. The driver can only run one path in a step. ...\\
Here is an example problem and the correct result. \\
... \\
Now please give me the result of the new planning problem driver-1 below, the solution's format should be the same as the example solution: \\
Given the planning problem driver-1 \\
(define (problem driver-1)
(:cities A B C D E F G)
(:constraints 
(You should have been to C and D before you go to G))
(:init
(road A B)
(road A E)
(road E D)
(road B C)
(road B F)
(road F G)
(reached A)
(at A)
)
(:goal
(and
(reached F)
(reached G))
))
\end{tcolorbox}

\begin{tcbitemize}[raster columns=2, before title=\vspace{0mm}, after title=\vspace{0mm}, raster equal height, top=0mm, bottom=-2mm]
\tcbitem[title= Maps, before title=\vspace{-1mm}, after title=\vspace{-1mm}]
\includegraphics[width=\linewidth]{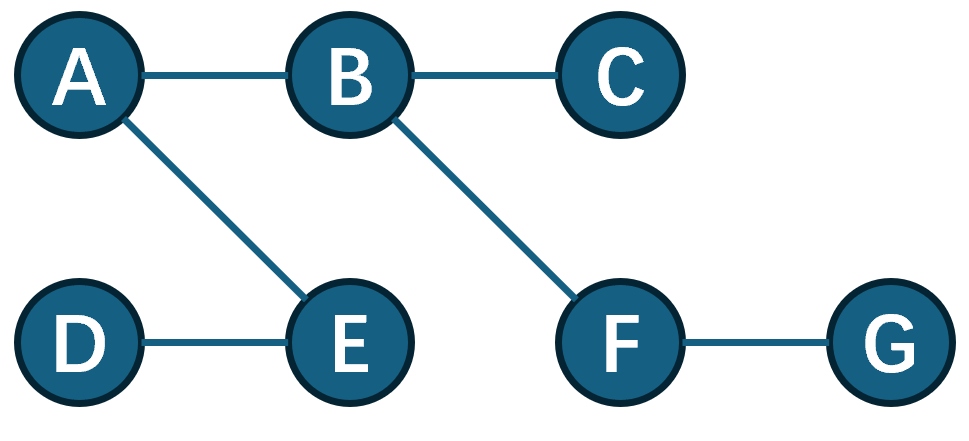}
\label{fig:map}

\tcbitem[title= 2\nd Output, before title=\vspace{-1mm}, after title=\vspace{-1mm}]
A $\rightarrow$ B \\
B $\rightarrow$ C \\
C $\rightarrow$ B \\
B $\rightarrow$ F \\
F $\rightarrow$ G \\
\end{tcbitemize}

\begin{tcbitemize}[raster columns=2, raster equal height, top=0mm, bottom=0mm]
\tcbitem[title= 3\rd Prompt, before title=\vspace{-1mm}, after title=\vspace{-1mm}]
A $\rightarrow$ B \\
B $\rightarrow$ C \\
C $\rightarrow$ B \\
B $\rightarrow$ F \\
F $\rightarrow$ G \\
is invalid. \\
It violates the constraint: You should have been to C and D before you go to G. \\

\tcbitem[title= 3\rd Output, sidebyside, before title=\vspace{-1mm}, after title=\vspace{-1mm}]
A $\rightarrow$ B \\ 
B $\rightarrow$ C \\
C $\rightarrow$ B \\
B $\rightarrow$ A \\
A $\rightarrow$ E \\
E $\rightarrow$ D \\
\tcblower
D $\rightarrow$ E \\
E $\rightarrow$ A \\
A $\rightarrow$ B \\
B $\rightarrow$ F \\
F $\rightarrow$ G \\
\end{tcbitemize}

The framework then takes these two prompt files and initiates its execution.
It first produces the LTL formula as the 1\st output, which, upon user validation of its accuracy.
The LTL formula for this scenario is expressed as: $\mathbf{G}(\neg g \, \mathcal{U} \, (c \land d))$,
where g, c, and d indicate the proposition that the driver has already reached cities G, C, or D, respectively.
This LTL formula format is supported by the verification tool Spot, which we use in our framework.
This LTL formula is then been transferred to an automaton in the formal specification component.

The LLM then begins to generate plan candidates.
We receive the 2\nd output from the LLM as detailed above.
After verification, our framework determines that the plan is invalid and provides reasoning as the 2\nd prompt.
The plan is deemed invalid because it directly proceeded to city G without passing through city D at the fifth step, thereby violating the temporal constraints.

The framework re-queries LLM using the 3\rd prompt and then gets the 3\rd output.
This plan passes all verifications within the framework and is output as the final correct result to the user or agent.
This experiment demonstrates that our method can obtain the correct plan from LLM while ensuring temporal constraints.

\subsubsection{Quantitative Experiments}
\begin{figure}[h]
    \centering   
    \includegraphics[width=\columnwidth]{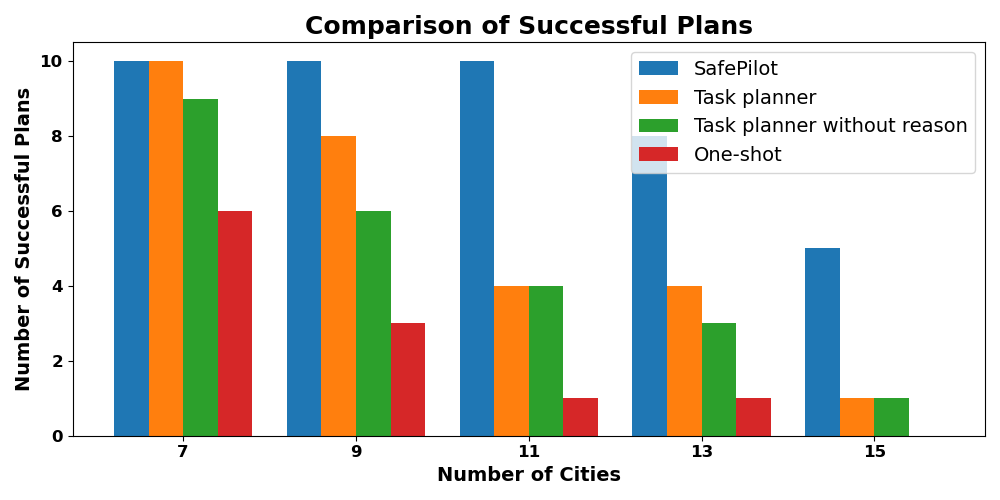}
    \caption{The performance of GPT-4o on Navigation problem. The blue, orange, and green bars represent the SafePilot, the task planner, the task planner without reason and one-shot methods, respectively.}
    \label{fig:nav}
\end{figure}
We evaluate the performance of our framework and baseline methods on 50 randomly generated navigation problems, varying in specifications.
In these problems, the number of cities is 7, 9, 11, 13 and 15, with ten problems for each quantity.
We select the same three baseline methods used in the experiments from the Blocksworld case study.
The results are shown in Fig.~\ref{fig:nav}.
The success rate of SafePilot is significantly higher than that of the three baselines.
This demonstrates that our framework can effectively handle problems with temporal requirements.

\section{Conclusion}~\label{conclusion}
We proposed SafePilot, a hierarchical neuro-symbolic framework for assuring LLM-enabled CPS.
Users can customize their tasks inheriting the built-in classes.
Moreover, we provide comprehensive experiments and examples of how to verify and guide LLM-enabled CPS to provide correct results.
In the future, we will integrate more tasks into our framework and support additional types of formal logic.
We will further enhance the planning capabilities of the LLM to reduce the number of iterations by developing better reasoning and fine-tuning the LLM.
Additionally, we will enhance the LLM's ability to generate logic formulas.
This will reduce the frequency of human expert reviews for LLM-generated logic formulas, thereby minimizing human effort.


\bibliographystyle{ACM-Reference-Format}
\bibliography{sample-base}

\newpage

\end{document}